\documentclass{article}
\usepackage{times}
\usepackage[pdftex]{graphicx}
\usepackage{multicol}
\usepackage{hyperref}
\usepackage{url}
\hypersetup{breaklinks=true}

\title{Neural Style Representations and the Large-Scale Classification of Artistic Style}

\author{Jeremiah W. Johnson \\
Department of Applied Sciences \& Engineering \\
University of New Hampshire \\
Manchester, NH 03101 \\
\texttt{jeremiah.johnson@unh.edu}}

\begin{document}

\maketitle

\begin{abstract}
The artistic style of a painting is a subtle aesthetic judgment used by art historians for grouping and classifying artwork. The recently introduced `neural-style' algorithm substantially succeeds in merging the perceived artistic style of one image or set of images with the perceived content of another. In light of this and other recent developments in image analysis via convolutional neural networks, we investigate the effectiveness of a `neural-style' representation for classifying the artistic style of paintings. \end{abstract}

\section{Introduction}

Any observer can sense the artistic style of painting, even if it takes training to articulate it. To an art historian, the artistic style is the primary means of classifying the painting \cite{Lang}. However, artistic style is not well defined, and may be loosely described as ``.. a distinctive manner which permits the grouping of works into related categories" \cite{Fernie}. Algorithmically determining the artistic style of an artwork is a challenging problem which may include analysis of features such as the painting's color, its texture, and its subject matter, or none of those at all. Detecting the style of a digitized image of a painting poses additional challenges raised by the digitization process, which itself has consequences that may affect the ability of a machine to correctly detect artistic style; for instance, textures may be affected by the resolution of the digitization. Despite these challenges, intelligent systems for detecting artistic style would be useful for identification and retrieval of images of a similar style.

In this paper we investigate several methods based on recent advances in convolutional neural networks for large-scale determination of artistic style. In particular, we adapt the neural-style algorithm introduced in \cite{gatys2016image} for large-scale style classification, showing performance that is competitive with other deep convolutional neural network based approaches.  

\begin{figure}[h]\label{starry}
\begin{center}
    \includegraphics[width=\linewidth]{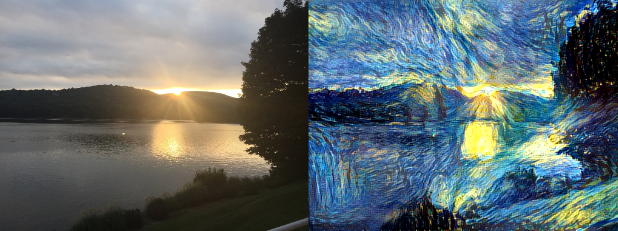}
\end{center}
\caption{Original image on the left, after application of the `neural-style' algorithm (style image 'Starry Night', by Van Gogh) on the right.}
\end{figure}  

\section{Related Work}

Algorithmic determination of artistic style in paintings has only been considered sporadically in the past. Examples of early efforts at style classification are \cite{Keren:2003:RIX:961320.961325} and \cite{Shamir:2010:IES:1670671.1670672}, where the datasets used are quite small, and only a handful of very distinct artistic style categories considered. Several complex models are constructed in \cite{saleh2015large} by hand-engineering features on a large dataset similar to the one used for this work. And in \cite{DBLP:journals/corr/KarayevHWAD13}, it is demonstrated that convolutional neural networks may be effective for understanding image style in general, including artistic style in paintings. In the papers just mentioned the number of artistic style categories is held to a relatively small 25 and 27 broadly defined style categories arespectively.   

In the paper ``A Neural Algorithm of Artistic Style'', it is demonstrated that the correlations between the low-level feature activations in a deep convolutional neural network encode sufficient information about the style of the input image to permit a tranfer of the visual style of the input image onto a new image via an algorithm informally referred to as the ``neural-style'' algorithm \cite{gatys2016image}. An example of the output of this algorithm is presented in Figure \ref{starry}. Several authors have built upon the work of Gatys et. al. in the past year \cite{ruder2016artistic}, \cite{DBLP:journals/corr/NovakN16}, \cite{DBLP:journals/corr/NovakN16}, \cite{Johnson2016Perceptual}. These investigations have primarily focused on ways to improve either the quality of the style transfer or the efficiency of the algorithm. To the best of our knowledge the only other look at the use of the style representation of an image as a classifier is in \cite{Matsuo:2016:CSV:2911996.2912057}. 

\section{Data and Methods}\label{methods}

\subsection{Data}
The data used for this investigation consists of 76449 digitized images of fine art paintings. The vast majority of the images were originally obtained from \url{http://www.wikiart.org}, the largest online repository of fine-art paintings. For convenience, we utilize a prepackaged set of images sourced and prepared by Kiri Nichols and hosted by the data-science competition website \url{http://www.kaggle.com}. A stratified 10\% of the dataset was held out for validation purposes. We chose to use a finer set of style categories for classification than has been used in previous work on image style, as we believe that finer classification is likely necessary for practical application. We utilize 70 distinct style categories, the maximum amount possible while maintaining at least 100 observations of each style category. This noticably increases the complexity of the classification task as many of the class boundaries are not well-defined, the classes are unbalanced, and there are not nearly as many examples of each of the artistic styles as in previous attempts at large-scale artistic style classification. 

\subsection{The Neural Style Algorithm}
The primary insight in the neural-style algorithm outlined by Gatys et. al. is that the correlations between low-level feature activations in a convolutional neural network capture information about the style of the image, while higher-level feature activations capture information about the content of the image. Thus, to construct an image $\mathbf{x}$ that merges both the style of an image $\mathbf{a}$ and the content of an image $\mathbf{p}$, an image is initialized as white noise and the following two loss functions are simultaneously minimized: 
\begin{equation}
    \mathcal{L}_{content}(\mathbf{p}, \mathbf{x}) = \sum_{l \in L_{content}} \frac{1}{N_lM_l}\sum_{i, j}\left(F_{ij}^l - P_{ij}^l\right)^2,
\end{equation}
and
\begin{equation}
    \mathcal{L}_{style}(\mathbf{a}, \mathbf{x}) = \sum_{l \in L_{style}} \frac{1}{N_l^2M_l^2}\sum_{i, j}(G_{ij}^l - A_{ij}^l)^2,
\end{equation}
where $N_l$ is the number of filters in the layer, $M_l$ is the spatial dimensionality of the feature map, $\mathbf{F}^l$ and $\mathbf{P}^l$ represent the feature maps extracted by the network at layer $l$ from the images $\mathbf{x}$ and $\mathbf{p}$ respectively, and letting $\mathbf{S}^l$ represent the feature maps extracted by the network at layer $l$ from the image $\mathbf{a}$, $\displaystyle G_{ij}^l = \sum_{k=1}^{M_l}F_{ik}^lF_{jk}^l$ and $\displaystyle A_{ij}^l = \sum_{k=1}^{M_l}S_{ik}^lS_{jk}^l$. That is, the style loss, which encodes the images style, is a loss taken over Gram matrices for filter activations.

\subsection{Style Classification}

\begin{table}[t]
\caption{Baseline Results}
\label{results}
\begin{center}
\begin{tabular}{ll}

\multicolumn{1}{c}{Model}  &\multicolumn{1}{c}{Accuracy (top 1\%)}
\\ \hline \\
Convolutional Neural Network        & 27.47 \\
Pretrained Residual Neural Network             & 36.99 \\
\end{tabular}
\end{center}
\end{table}

To establish a baseline for style classification, we first trained a single convolutional neural network from scratch. The network has a uniform structure consisting of convolutional layers with 3x3 kernels and leaky ReLUs activations ($\alpha = 0.333$). Between every pair of convolutional layers is a fractional max pooling layer with a 3x3 kernel. Fractional max-pooling is used as given the relatively small size of the dataset, the more commonly used average or max-pooling operations would lead to rapid data loss and a relatively shallow network \cite{DBLP:journals/corr/Graham14a}. The convolutional layer sizes are $3 \to 32 \to 96 \to 128 \to160\to 192 \to 224,$ followed by a fully-connected layer and 70-way softmax. 10\% dropout is applied to the fully connected layer. Aside from mean normalization and horizontal flips, the data were not augmented in any way. The model was trained over 55 epochs using stochastic gradient descent and achieved a top 1\% accuracy of 27.468\%.

We then finetuned a pretrained object classification model for style classification. The pretrained model used was a residual neural network with 50 layers pretrained on the ImageNet 2015 dataset. There are two motivating factors for choosing to finetune this network. The first is that residual networks currently exhibit the best on object recognition tasks, and previous work on style classification suggests that a network trained for the task of object recognition and then finetuned for image style detection will perform the task well \cite{DBLP:journals/corr/HeZRS15}, \cite{DBLP:journals/corr/KarayevHWAD13}. The second and more interesting reason from the standpoint of artistic style classification is that the architecture of a residual neural network makes the outputs of lower levels of the network available to higher levels in the network. In this way, the net functions similar to a Long Short-Term Memory network without gates \cite{srivastava2015training}. For style classification, this is particularly appealing as a means of allowing the higher levels in the net to consider both lower-level features and higher-level features when forming an artistic style classification, where the style may very much be determined by the lower-level features. The residual neural network model obtained top-1\% accuracy of 36.985\%.  

To determine whether or not the style representation encoded in the Gram matrices for a given image has any power as a classifier, we extracted the Gram matrices of feature activations at layers ReLU1\_1, ReLU2\_1, ReLu3\_1, ReLu4\_1, and ReLU5\_1 from a VGG-19 network for the paintings described above \cite{DBLP:journals/corr/SimonyanZ14a}. The choice of network and layers was based on the quality of the style transfers obtained with these choices in \cite{gatys2016image}. The pretrained VGG-19 model was obtained from the Caffe Model Zoo \cite{jia2014caffe}. The Gram matrices were then reshaped to account for symmetry, producing a total of 304,416 distinct features per image, nearly a factor of four greater that the total number of observations in the dataset. 

Analyzing the style representation was approached in two ways. First, the full feature vector was normalized and then passed to a single-layer linear classifier which was trained using Adam over 55 epochs, producing a top 1\% accuracy of 13.23\% \cite{kingma2014adam}.  

We then built random forest classifiers on the individual Gram matrices extracted from the activations of the network. The dimensionality of the Gram matrices post-reshaping is 2016, 8128, 32640, 130816, and 130816 respectively. Considered separately, the random forest classifiers built on the first three of these style representations performed better than the linear classifier based on the full style representation and better than the baseline convolutional neural network, with top-1\% accuracies of 27.84\%, 28.97\%, and 33.46\%. The random forests built on the latter two layers performed considerably worse. The results are presented in table \ref{style_results}. 

In contrast to results reported in \cite{Matsuo:2016:CSV:2911996.2912057}, we observed a significant loss in accuracy when dimensionality reduction was even lightly utilized on these smaller layers. For instance, performing PCA while preserving 90\% of the variance in the data from the layer ReLU1\_1 style representation reduced the accuracy of the random forest model on that layer from 27.84\% to 17\%, perhaps due to our use of a larger, less homogeneous dataset. We also saw no significant gains when the data were normalized.
 
\begin{table}[t]\label{style-results}
\caption{Style Representation Results}
\label{style_results}
\begin{center}
\begin{tabular}{ll}

\multicolumn{1}{c}{Model}  &\multicolumn{1}{c}{Accuracy (top 1\%)}
\\ \hline \\
Full Style Representation - Linear Classifier               & 13.21 \\
ReLU1\_1 Random Forest                  & 27.84 \\
ReLU2\_1 Random Forest                  & 28.97 \\
ReLU3\_1 Random Forest                  & 33.46 \\
ReLU4\_1 Random Forest                  & 9.79 \\
ReLU5\_1 Random forest                  & 10.18 
\end{tabular}
\end{center}
\end{table}

\section{Conclusion \& Future Work}
 The `neural-style' representation of an artwork offers competitive performance as an artistic style classifier; nevertheless, in our experiments a finetuned deep neural network still obtains superior results. Our best results using the `neural-style' representation of artistic style were obtained when models suitable for high-dimensional nonlinear data were constructed individually on the first three Gram matrices that form the building blocks of the style representation. 

It appears that the art-historical definition of artistic style is not quite what is captured by the neural style algorithm using this network and these layers. Nevertheless it is clear that this information is relevant and has some predictive ability, and understanding and improving on these results is a target for future work.


\subsubsection*{Acknowledgments}
The author would like to thank NVIDIA for GPU donation to support this research, Wikiart.org for providing many of the images, the website Kaggle.com for hosting the data, and Kiri Nichols for sourcing the data.
\bibliography{artbib}
\bibliographystyle{acm}
\newpage

\subsubsection*{Supplementary Material}

To create the following visualizations, 20000 images were selected at random and their Gram matrices of activations at layers ReLU1\_1, ReLU2\_1, and ReLU3\_1 were extracted. The visualizations were produced by then running the Barnes-Hut $t$-SNE algorithm on the Gram matrices at each layer, rather than on the raw pixel data.
\begin{figure}[h]\label{relu1_1_tsne}
\begin{center}
    \includegraphics[width=\linewidth]{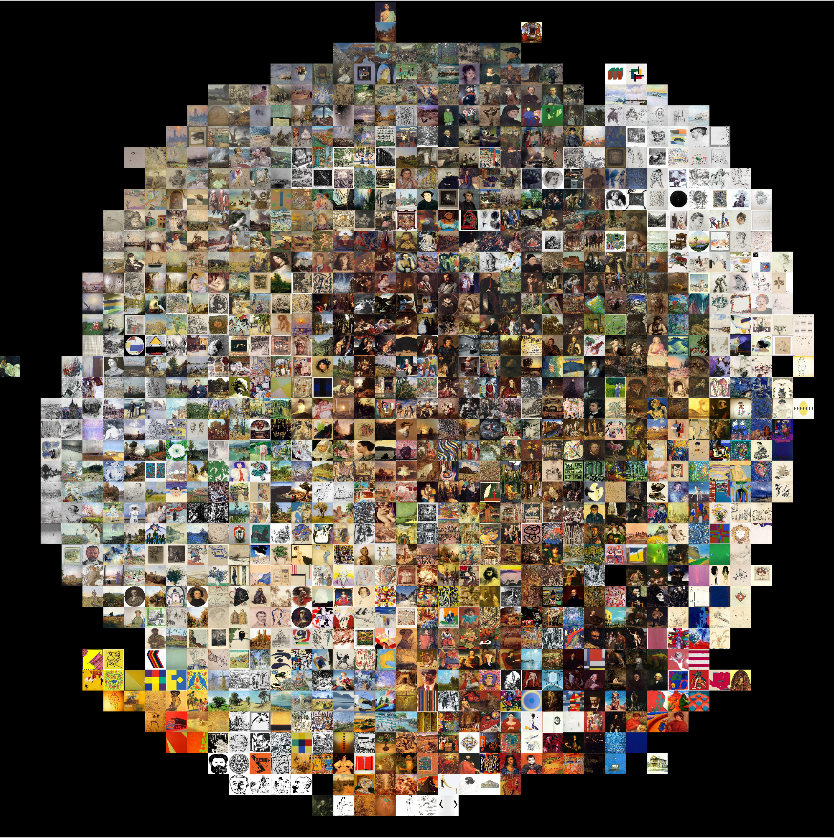}
\end{center}
\caption{A Barnes-Hut $t$-SNE visualization of 20000 randomly selected images. The features used in creating the visualization are the components of the Gram matrix from layer ReLU1\_1.}
\end{figure}

\begin{figure}[h]\label{relu2_1_tsne}
\begin{center}
    \includegraphics[width=\linewidth]{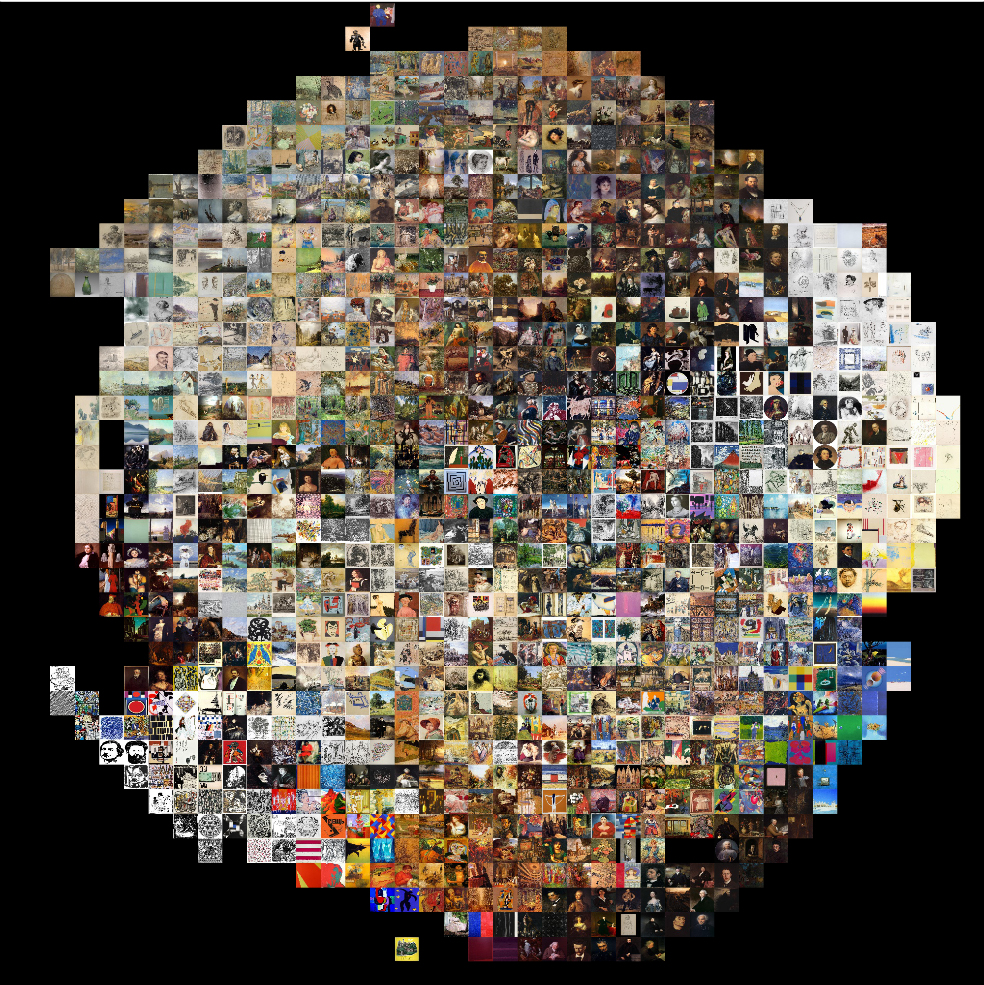}
\end{center}
\caption{A Barnes-Hut $t$-SNE visualization of 20000 randomly selected images. The features used in creating the visualization are the components of the Gram matrix from layer ReLU2\_1.}
\end{figure}

\begin{figure}[h]\label{relu3_1_tsne}
\begin{center}
    \includegraphics[width=\linewidth]{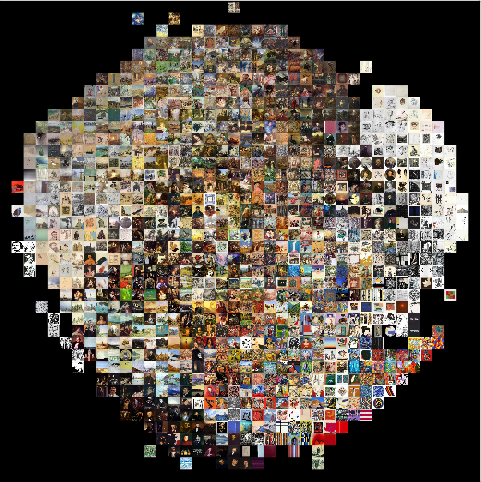}
\end{center}
\caption{A Barnes-Hut $t$-SNE visualization of 20000 randomly selected images. The features used in creating the visualization are the components of the Gram matrix from layer ReLU3\_1.}
\end{figure}

\end{document}